\documentclass[letterpaper, 10 pt, conference]{ieeeconf}  

\IEEEoverridecommandlockouts                              

\usepackage{cite}
\usepackage{amsmath,amssymb,amsfonts}
\usepackage{algorithmic}
\usepackage{graphicx}
\usepackage{textcomp}
\usepackage{xcolor}
\usepackage{multirow}
\usepackage{threeparttable}
\usepackage{array}
\usepackage{booktabs}
\usepackage{amsbsy}
\usepackage{marvosym}
\usepackage[colorlinks,bookmarksopen,bookmarksnumbered,citecolor=green, linkcolor=red, urlcolor={blue!50}]{hyperref}

\title{\LARGE \bf
Uni-Gaussians: Unifying Camera and Lidar Simulation with Gaussians for Dynamic Driving Scenarios
}

\author{Zikang~Yuan$^{1*}$, Yuechuan~Pu$^{2*}$, Hongcheng~Luo$^{2*}$, Fengtian~Lang$^{3}$, Cheng~Chi$^{2}$, Teng~Li$^{2}$,\\ Yingying~Shen$^{2}$, Haiyang~Sun$^{2\dagger}$, Bing~Wang$^{2}$ and Xin~Yang$^{3}$\textsuperscript{\Letter} \\ \url{https://zikangyuan.github.io/UniGaussians/}
	\thanks{$^{1}$Zikang~Yuan is with AI Chip Center for Emerging Smart Systems, InnoHK Centers, Hong Kong Science Park, Hong Kong, China. (E-mail: {\tt\small zikangyuan@ust.hl})}%
	\thanks{$^{2}$Yuechuan~Pu, Hongcheng~Luo, Cheng~Chi, Teng~Li, Yingying~Shen, Haiyang~Sun and Bing~Wang are with Xiaome EV, China. (E-mail: {\tt\small luohongcheng@xiaomi.com})}%
	\thanks{$^{3}$Fengtian~Lang and Xin~Yang are with the Electronic Information and Communications, Huazhong University of Science and Technology, Wuhan, 430074, China. (E-mail: {\tt\small xinyang2014@hust.edu.cn})}%
	\thanks{$^*$ means that these authors contributed equally to this work; $^\dagger$ means the project leader; $\textsuperscript{\Letter}$ means the corresponding author.}
}

\begin{document}

\maketitle
\thispagestyle{empty}
\pagestyle{empty}

\begin{abstract}

Ensuring the safety of autonomous vehicles necessitates comprehensive simulation of multi-sensor data, encompassing inputs from both cameras and LiDAR sensors, across various dynamic driving scenarios. Neural rendering techniques, which utilize collected raw sensor data to simulate these dynamic environments, have emerged as a leading methodology. While NeRF-based approaches can uniformly represent scenes for rendering data from both camera and LiDAR, they are hindered by slow rendering speeds due to dense sampling. Conversely, Gaussian Splatting-based methods employ Gaussian primitives for scene representation and achieve rapid rendering through rasterization. However, these rasterization-based techniques struggle to accurately model non-linear optical sensors. This limitation restricts their applicability to sensors beyond pinhole cameras. To address these challenges and enable unified representation of dynamic driving scenarios using Gaussian primitives, this study proposes a novel hybrid approach. Our method utilizes rasterization for rendering image data while employing Gaussian ray-tracing for LiDAR data rendering. Experimental results on public datasets demonstrate that our approach outperforms current state-of-the-art methods. This work presents a unified and efficient solution for realistic simulation of camera and LiDAR data in autonomous driving scenarios using Gaussian primitives, offering significant advancements in both rendering quality and computational efficiency\footnote{https://zikangyuan.github.io/UniGaussians/}.

\end{abstract}

\section{Introduction}
\label{sec:intro}

To ensure the safety of mobile robots and autonomous vehicles \cite{yuan2023sdv, yuan2022sr, yuan2023semi, yuan2024sr, yuan2023liwo}, extensive simulations of autonomous driving scenarios are required. Current autonomous vehicles are typically equipped with multiple cameras and a 3D spinning LiDAR. To more comprehensively and accurately replicate real-world scenarios, it is necessary to conduct simulation of both camera and LiDAR mounted data. Neural rendering methods have garnered significant attention in the field of simulation for autonomous driving, as they can construct simulation environments from collected logs in a data-driven manner.

\begin{figure*}[t]
	\centering
	\includegraphics[width=1.0\linewidth]{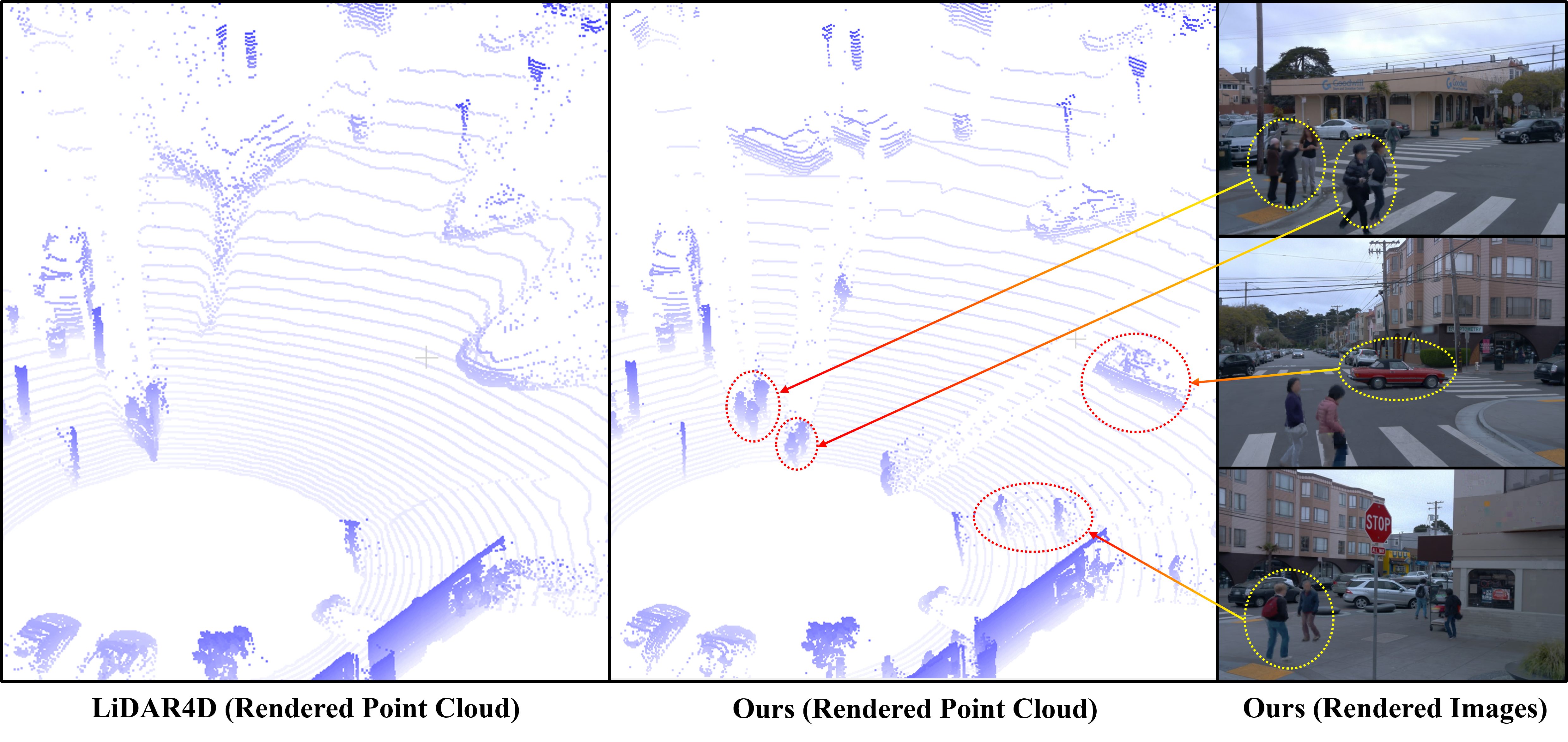}
	
	\caption{Point cloud simulation results of the newest SOTA method LiDAR4D and Ours. Our method can accurately simulate various movable entities including both rigid vehicles and non-rigid pedestrians while LiDAR4D fails.}
	\label{fig:first}
\end{figure*}

Existing representative neural rendering methods are mainly divided into neural radiance field (NeRF) and gaussian splatting (GS). NeRF-based methods offer high-fidelity simulations for both camera and LiDAR data, aligning with the prevalent sensor configurations used in autonomous driving. However, the ray-tracing rendering for dense sampling of NeRF-based methods results in slow rendering speeds, making them costly and challenging to be used for large-scale evaluation and analysis. GS-based methods serves as a viable alternative to NeRFs, achieving comparable levels of image realism while significantly enhancing rendering efficiency. Different from NeRF-based methods, GS-based methods represent the scenarios as Gaussian primitives, and utilize rasterization to achieve fast rendering. However, the conventional rasterization approach employed in Gaussian Splatting \cite{kerbl20233d} is not compatible with the rendering of cylindrical range images produced by LiDAR sensors. Moreover, the rasterization imaging process does not align with the active sensing mechanism of LiDAR sensors, which involves the emission and reception of laser beams. This precludes the Gaussian primitives from being uniformly adapted to the simulation tasks of LiDAR point cloud despite the fast rendering speed of GS-based methods.

In this paper, we propose a unifying camera and LiDAR simulation with Gaussians for dynamic driving scenarios. For unified representing the dynamic driving scenarios using Gaussian primitives, it is essential to select distinct rendering methods in accordance with the characteristics of different sensor data. Currently, the common rendering methods for Gaussian-based scenarios include rasterization and ray-tracing. For image data, the prior work \cite{moenne20243d} has demonstrated that the rendering quality of ray-tracing is comparable to that of rasterization, but at the cost of significantly higher computational time. Therefore, we utilize rasterization to render image data. For LiDAR data, the rasterization imaging process does not align with the active sensing mechanism of LiDAR sensors. Consequently, we select ray-tracing to render LiDAR data. Refer to OmniRe \cite{chen2024omnire}, we also perform graph modeling of the entire dynamic Gaussian scenarios (e.g., background, rigid vehicles and non-rigid pedestrians). Thus, we are capable of conducting comprehensive simulation of all elements within the entire Gaussian dynamic scenarios for both camera and LiDAR data. Experimental results on the public dataset demonstrate that our method can achieve comparable or more outstanding rendering performance than state-of-the-art (SOTA) methods (as shown in Fig. \ref{fig:first}), and our method offers a unified and efficient solution for realistic simulation from camera and LiDAR data of autonomous driving scenarios with Gaussian primitives.

To summarize, the main contributions of this work are three folds: 1) We propose a unified and efficient solution for realistic simulation from camera and LiDAR data with Gaussians; 2) We achieve the high-fidelity LiDAR simulation that encompasses all dynamic actors including vehicles, pedestrians and cyclists; 3) We perform extensive experiments and ablations to demonstrate the benefits of utilizing our unified Gaussian representation and separate rendering approaches.

The rest of this paper is structured as follows. Sec. \ref{sec:related} reviews the relevant literature. Secs. \ref{sec:method} details the methodology, followed by experimental evaluation in Sec. \ref{sec:experiment}. Sec. \ref{sec:conclusion} concludes the paper.

\section{Related Work}
\label{sec:related}

Neural representations have become predominant in the domain of novel view synthesis \cite{mildenhall2021nerf, barron2022mip, barron2021mip, muller2022instant, fridovich2022plenoxels, kerbl20233d}. Various extensions of these representations have been developed to facilitate the simulation of dynamic scenes. This section will provide a discussion of the three mainstream methods, NeRF, GS and Gaussian ray-tracing (GRT).

\textbf{NeRF-based methods.} In recent years, numerous studies based on neural radiance fields \cite{barron2021mip, chan2022efficient, chen2022tensorf, fridovich2022plenoxels, hu2023tri, liu2020neural, mildenhall2021nerf, muller2022instant, sun2022direct} have achieved breakthrough progress and significant accomplishments in the task of novel view synthesis (NVS). Various neural representations based on MLPs \cite{barron2021mip, mildenhall2021nerf}, voxel grids \cite{fridovich2022plenoxels, liu2020neural, sun2022direct}, triplanes \cite{fridovich2022plenoxels, hu2023tri}, vector decomposition \cite{chen2022tensorf}, and multilevel hash grids \cite{muller2022instant} have been fully utilized for reconstruction and synthesis. Several recent works \cite{barron2022mip, rematas2022urban, wang2023neural} have even gradually expanded their scale from object-centered indoor small scenes to large outdoor scenes. However, due to the rendering method of dense samples, which requires rendering densely sampled points along a ray, the rendering overhead of NeRF is substantial, making them costly and challenging to be used for large-scale evaluation and analysis. On the other hand, the ray-tracing rendering method is well-suited for sensors that acquire sparse point clouds through ray sampling, such as LiDAR. In recent years, several NeRF-based methods utilizing LiDAR data \cite{zheng2024lidar4d, huang2023neural, tao2024lidar, yang2023unisim, zhang2024nerf} have been proposed. Among LiDAR-only methods, LiDAR-NeRF \cite{tao2024lidar} and NFL \cite{huang2023neural} propose a differentiable LiDAR novel view synthesis (NVS) framework for static scenes, simultaneously reconstructing depth, intensity, and ray-drop probability. UniSim \cite{yang2023unisim} builds neural
feature grids to reconstruct both the static background and dynamic actors in the scene, and composites them together to simulate LiDAR and camera data at new viewpoints. LiDAR4D \cite{zheng2024lidar4d} proposes a 4D hybrid representation that combines multi-plane and grid features, enabling effective coarse-to-fine reconstruction. Similar to the processing of image data, LiDAR-based methods also perform dense sampling along each laser ray and subsequently render the sampled points to realistically synthesize LiDAR point clouds. However, this ray-tracing rendering approach for dense sampling requires substantial computational time, making large-scale testing challenging.

\textbf{GS-based methods.} Gaussian Splating \cite{kerbl20233d} defines a set of anisotropic Gaussians in the 3D world and performs adaptive density control to achieve high-quality rendering results from sparse point clouds. Recent methods have extended 3D GS to small-scale dynamic scenes by introducing deformation fields \cite{wu20244d, yang2024deformable}, physical priors \cite{luiten2024dynamic}, or 4D parameterization \cite{yang2023real} into the 3D Gaussian model. Recently, several concurrent studies have also explored 3D Gaussians in urban street scenes. DrivingGaussian \cite{zhou2024drivinggaussian} introduces incremental 3D static Gaussian maps and composite dynamic Gaussian maps. PVG \cite{chen2023periodic} utilizes a periodic vibration 3D Gaussian model to simulate dynamic urban scenes. In StreetGaussian \cite{yan2024street}, dynamic urban scenes are represented as a set of point clouds equipped with semantic logic and 3D Gaussian functions, each associated with either foreground vehicles or background. To simulate the dynamics of foreground objects (vehicles), each object point cloud is optimized using an optimizable tracking pose and a 4D spherical harmonics model for dynamic appearance. OmniRes \cite{chen2024omnire} first achieved the simulation of non-rigid dynamic objects in dynamic driving scenes. It constructs a dynamic neural scene graph based on Gaussian representation and builds multiple local canonical spaces to model various types of dynamic objects, including vehicles, pedestrians, and cyclists. Unlike ray-tracing, the rasterization offers faster rendering speed, which makes it a more preferred choice for large-scale simulations. SplatAD \cite{hess2024splatad} and LiDAR-GS \cite{chen2024lidar} try to utilize rasterization to render LiDAR data. However, the imaging process of rasterization is not consistent with the principle of data acquisition employed by LiDAR sensors. Therefore, the simulation results achieved using rasterization for LiDAR data are unsatisfactory.

\textbf{GRT-based methods.} In recent years, 3DGRT-based methos \cite{gu2024irgs, moenne20243d, xie2024envgs} propose to calculate the intersection of Gaussian ellipsoids and a given ray, and then render the Gaussian primitives involved with these intersections. They point out that compared to rasterization, Gaussian ray tracing can better simulate the visual effects of environmental lighting. In this work, we demonstrate that Gaussian ray tracing can effectively address the issue of inaccurate LiDAR data rendering caused by Gaussian rasterization. Using rasterization for image data and Gaussian ray tracing for LiDAR data enables effective unified rendering of Gaussian scenarios.

\section{Methodlogy}
\label{sec:method}

\subsection{Overview}
\label{sec:overview}

\begin{figure*}[t]
	\centering
	\includegraphics[width=1.0\linewidth]{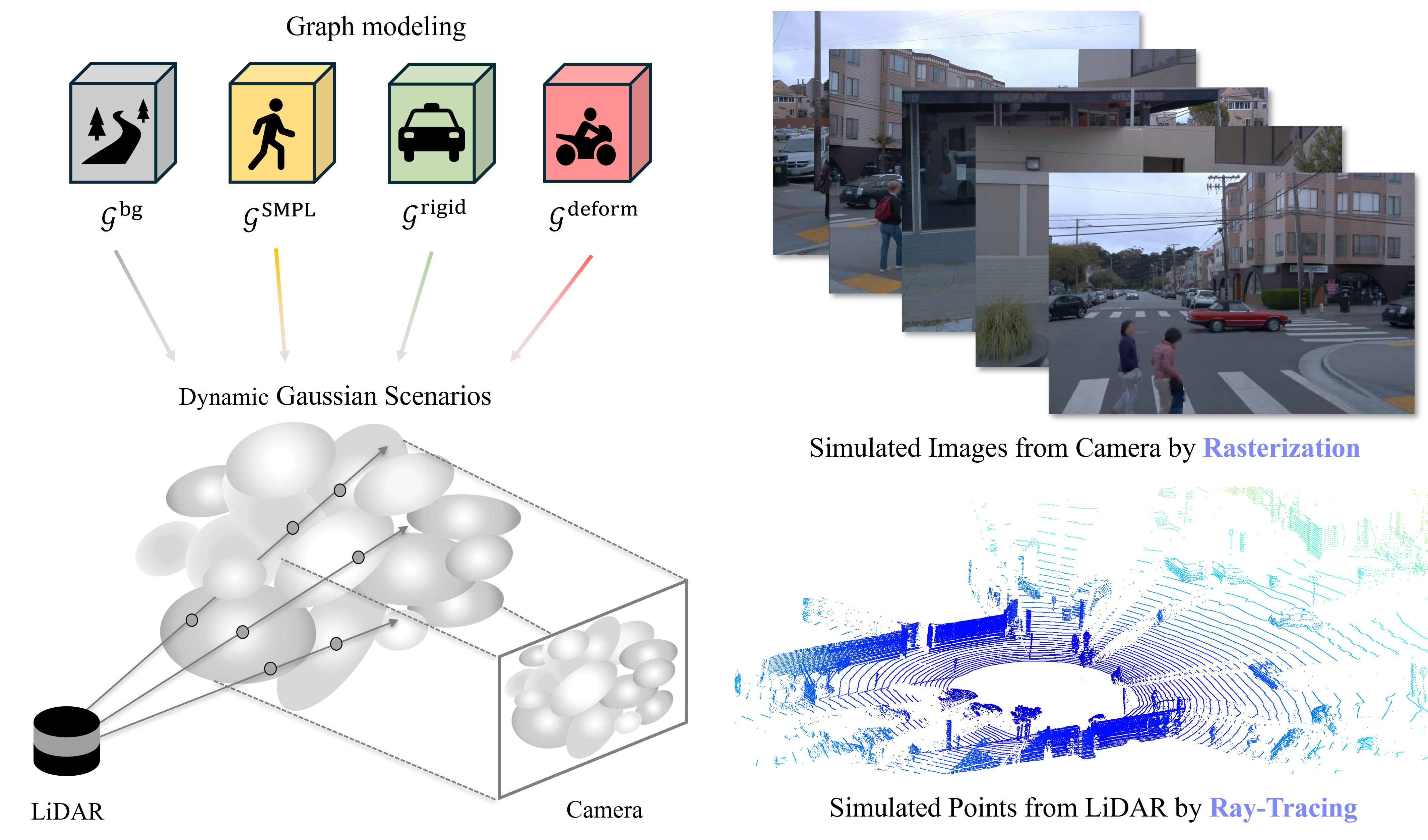}
	
	\caption{Method Overview. Gaussians of all elements are defined in their local or canonical spaces, and are deformed and transformed into the world space at a given time $t$. We perform unifying camera and LiDAR simulation for the whole dynamic driving scenarios. For camera image data, we employ rasterization for rendering. For LiDAR data, we compute the intersection between ellipses and rays to construct ray-tracing.}
	\label{fig:framework}
\end{figure*}

Fig. \ref{fig:framework} illustrates the overview of our unifying camera and LiDAR simulation framework. Inspired by OmniRe \cite{chen2024omnire}, we establish a Gaussian scene graph representation, which encompasses both the static background and various movable entities, including rigid vehicles and non-rigid pedestrians. After graph modeling, for a given time $t$, we can deform and transform the Gaussian primitives of the movable entities into the world space, where they are combined with the background Gaussian primitives to form the Gaussian scene graph, thereby modeling the entire scenarios.

At present, the prevalent rendering approaches for Gaussian-based scenes are rasterization and ray-tracing. For regarding image data, previous research \cite{moenne20243d} has shown that ray-tracing achieves rendering quality on par with rasterization, albeit with a notable increase in computational time. Given this trade-off, rasterization is chosen for rendering image data. On the other hand, the rasterization method does not support the rendering of cylindrical range images produced by LiDAR sensors. Furthermore, the rasterization imaging process is not aligned with the active nature of LiDAR sensors, which rely on the emission and reception of laser beams. This limitation leads to point positional disorder in the simulation of point clouds, which are demonstrated in Sec. \ref{sec:ablation2}. As a result, ray-tracing is deemed more suitable and used for rendering LiDAR data. The detailed rendering methodologies will be detailed in Sec. \ref{sec:camera} and \ref{sec:lidar}.

\subsection{Modeling}
\label{sec:model}

In the application of 3D Gaussian Splatting (3D GS) for camera simulation methods, scene graph modeling has been extensively utilized. In the most recent camera simulation method OmniRe \cite{chen2024omnire}, dynamic driving scenarios are decomposed into static backgrounds and foreground objects, with the latter further classified into rigid vehicles and non-rigid pedestrians. However, such detailed and accurate modeling is conspicuously absent in LiDAR simulation. Inspired by OmniRe, we adopt an analogous scene graph approach to model the movable entities (the rigid node $\mathcal{G}^{\text {rigid }}$, the non-rigid SMPL node $\mathcal{G}^{\text {SMPL}}$ and the non-rigid deformable node $\mathcal{G}^{\text {deform}}$ within the scene for better camera-LiDAR simulation. On the other hand, the initial design of 3D Gaussians was primarily intended for realistic novel-view synthesis, whereas point cloud simulation emphasizes accurate geometric reconstruction. Specifically, the utilization of 3D Gaussians for geometric surface reconstruction faces several limitations: 1) The volumetric emission representation of 3D Gaussians conflicts with the thin nature of surfaces. 2) 3D Gaussians do not natively model surface normals, which can lead to surface artifacts in point cloud simulations. 3) The rasterization process in 3D GS lacks multi-view consistency, resulting in different 2D intersecting planes from various viewpoints. To better ensure the geometric representation for simulation, we employ 2D Gaussians \cite{huang20242d} to represent the entire dynamic driving scenario.

Similar as \cite{huang20242d}, we represent each 2D Gaussian primitive as
\begin{equation}
	\left\{\mathbf{p}, \mathbf{t}_u, \mathbf{t}_v, s_u, s_v, \alpha, c, i, r\right\}
	\label{eq:10},
\end{equation}
where $\mathbf{p}$ is the center point, $\mathbf{t}_u$ and $\mathbf{t}_v$ are the two tangent vectors, $s_u$ and $s_v$ are the scaling factor that controls the variances of 2D Gaussian distribution along $\mathbf{t}_u$ and $\mathbf{t}_v$, $\alpha$ is the opacity, $c$ is the Spherical Harmonic (SH) color, $i$ is the SH intensity and $r$ is the SH ray-drop probability. All of them are learnable parameters. $i$ and $r$ are additional attributes we introduce based on the original 2D Gaussian primitive to better adapt to the characteristics of LiDAR sensors. $i$ corresponds to the reflectivity of the laser ray, and $r$ is a binary variable indicating whether the data collected by the ray is valid. It is worth emphasizing that the utilization of SH coefficients is motivated by their ability to account for the influence of viewpoint variations on color. For LiDAR intensity and ray-drop probability which are also view dependent, we also use SH coefficients to express them for enhancing the realism of simulation.

\subsection{Camera Image Rendering}
\label{sec:camera}

We employ rasterization to render camera image data. As mentioned in \cite{zwicker2004perspective}, we express the projection of 2D splats onto the image plane using a 2D-to-2D mapping in homogeneous coordinates. Let $\mathbf{W}\in4\times4$ denote the composite transformation matrix from world space to screen space. The screen space points are represented by
\begin{equation}
	\mathbf{x}=(x z, y z, z, z)^T=\mathbf{W H}(u, v, 1,1)^T,
	\label{eq:1}
\end{equation}
\begin{equation}
	\mathbf{H}=\left[\begin{array}{cccc}
		s_u \mathbf{t}_u & s_v \mathbf{t}_v & \mathbf{0} & \mathbf{p} \\
		0 & 0 & 0 & 1
	\end{array}\right],
	\label{eq:2}
\end{equation}
where $\mathbf{x}$ denotes the uniform ray emanating from the camera, passing through the pixel $(x,y)$ and intersecting the splat at depth $z$. In practice, we efficiently determine the intersection of the ray and the splat by finding the intersection point $\mathbf{u}(\mathbf{x})$ of three non-parallel planes \cite{weyrich2007hardware}.

The entire rasterization process is analogous to that of 3D Gaussian Splats. Initially, a screen-space bounding box is computed for each Gaussian primitive. Subsequently, the 2D Gaussian primitives are sorted based on the depth of their centers and organized into tiles according to their bounding boxes. Finally, the alpha-weighted appearance is integrated from front to back using volumetric alpha blending:
\begin{equation}
	\mathbf{c}(x)=\sum_{j=1} \mathbf{c}_j \alpha_j {\mathcal{G}}_j(\mathbf{u}(\mathbf{x})) \prod_{k=1}^{j-1}\left(1-\alpha_k {\mathcal{G}}_k(\mathbf{u}(\mathbf{x}))\right),
	\label{eq:3}
\end{equation}
where $\mathbf{c}_j$ is the three-channel color mapped from the SH color of each Gaussian primitive and the projection direction. The iterative process is terminated when the accumulated opacity reaches saturation.

\subsection{LiDAR Point Cloud Rendering}
\label{sec:lidar}

As previously mentioned, the rasterization-based rendering approach is not directly applicable to LiDAR point cloud data. Therefore, to enable unifying camera-LiDAR rendering with Gaussian primitives, we employ Gaussian ray-tracing \cite{kerbl20233d, xie2024envgs} to render the LiDAR point cloud data.

To fully leverage the hardware acceleration of ray-primitive intersections, each 2D Gaussian primitive is transfprmed into a geometric primitive compatible with GPU processing and insert it into a bounding volume hierarchy (BVH). To this end, two triangles are utilized to represent 2D Gaussian distribution. After the transformation, the triangles are organized into a BVH, which is then used as input for the ray-tracing process.

Like existing GRT-based methods \cite{moenne20243d, xie2024envgs}, we develope custom CUDA kernels using the $raygen$
and $anyhit$ programmable entry points of OptiX \cite{parker2010optix}, which is also utilized in our framework. The rendering is performed in a block-by-block manner. The $anyhit$ kernel tracks the input rays to obtain blocks of size $k$, while the $raygen$ kernel integrates these blocks and invokes the $anyhit$ kernel to retrieve the next block along the ray. Specifically, the ray generation program first initiates traversal of the BVH to identify all potential intersections along the ray. During traversal, the $anyhit$ program sorts each intersected Gaussian by depth and maintains a sorted $k$-buffer for the $k$ nearest intersections. The $raygen$ program integrates the properties of the sorted Gaussian primitives in the buffer and computes the Gaussian values at the transformed points. This process is iteratively performed until no further intersections are detected along the ray or the accumulated transmittance falls below a specified threshold. For more details regarding the intersection of rays and 2D Gaussian primitives, please refer to \cite{moenne20243d, xie2024envgs}.

We define the set of ray-Gaussian intersections as $\{\mathbf{p}_1$, $\mathbf{p}_2$, $\cdots$, $\mathbf{p}_m\}$. Based on the distance from each intersection point to $\mathbf{r}_o$, we can calculate the distance $d_j$ for each intersection. Concurrently, by referencing the associated 2D Gaussian primitive of each intersection point, we can directly obtain the corresponding SH intensity $i_j$ and SH ray-drop probability $r_j$. Above SH coefficients and ray directions are input into the spherical harmonic functions to obtain the mapped intensity $\mathbf{i}_j$ and ray-drop probability $\mathbf{r}_j$. Finally, the alpha-weighted appearance is integrated from front to back using volumetric alpha blending:
\begin{equation}
	D,I,R=\sum_{j=1}^m v_j \alpha_j \mathcal{G}_j \prod_{k=1}^{j-1}\left(1-\alpha_k\mathcal{G}_k\right), v_j\in\{d_j, \mathbf{i}_j, \mathbf{r}_j\},
	\label{eq:7}
\end{equation}
where $\mathcal{G}_{.}$ is the standard 2D Gaussian value evaluation. The whole ray-tacing process is fully differentiable and the iterative process is terminated when the accumulated opacity reaches saturation.

\subsection{Optimization}
\label{sec:opt}

Our framework is fully differentiable, thereby enabling end-to-end optimization of both the base and environment Gaussian primitives. We optimize all learnable parameters mentioned in Sec. \ref{sec:model} to simulate the whole dynamic driving scenarios. These learnable parameters include all Gaussian properties (center $\mathbf{p}$, scale factor $s_u$ and $s_v$, rotation $\mathbf{t}_u$ and $\mathbf{t}_v$, opacity $\alpha$, color appearance $c$, intensity $e$, ray-drop probability $b$). We use the following loss for optimization:
\begin{equation}
	\begin{aligned}
		\mathcal{L}&=\left(1-\lambda_r\right) \mathcal{L}_1 + \lambda_r \mathcal{L}_{\text{SSIM}}+\lambda_{\text {depth}} \mathcal{L}_{\text{depth}} \\ 
		& +\lambda_{\text{intensity}} \mathcal{L}_{\text{intensity}} +\lambda_{\text{raydrop}} \mathcal{L}_{\text{raydrop}} \\
		& + \lambda_{\text{normal}} \mathcal{L}_{\text {normal}},
	\end{aligned}
	\label{eq:10}
\end{equation}
where $\mathcal{L}_1$ and $\mathcal{L}_{\text{SSIM}}$ are the $\mathcal{L}_1$ and SSIM losses on rendered images, $\mathcal{L}_{\text{depth}}$ is the $\mathcal{L}_1$ loss for comparing the rendered depth of Gaussians with depth signals from LiDAR, $\mathcal{L}_{\text{intensity}}$ is the $\mathcal{L}_2$ loss on rendered intensity, $\mathcal{L}_{\text{raydrop}}$ is the $\mathcal{L}_2$ loss on rendered ray-drop probability, $\mathcal{L}_{\text {normal }}$ is a normal consistency constraint \cite{huang20242d, xie2024envgs} between the rendered normal map and the gradients of the depth map.

\section{Experiments}
\label{sec:experiment}

\begin{table*}
	\centering
	\fontsize{7pt}{10pt}\selectfont
	\begin{tabular}{p{1.5cm}<{\centering}|p{0.6cm}<{\centering}p{0.9cm}<{\centering}|p{0.5cm}<{\centering}|p{0.4cm}<{\centering}|p{0.7cm}<{\centering}p{0.7cm}<{\centering}p{0.7cm}<{\centering}p{0.65cm}<{\centering}p{0.65cm}<{\centering}|p{0.7cm}<{\centering}p{0.7cm}<{\centering}p{0.7cm}<{\centering}p{0.65cm}<{\centering}p{0.65cm}<{\centering}}
		\hline
		\multirow{2}{*}{Method} & \multicolumn{2}{c|}{Point Cloud} & \multirow{2}{*}{\begin{tabular}[c]{@{}c@{}}Infer \\ Time$\downarrow$\end{tabular}} & \multirow{2}{*}{\begin{tabular}[c]{@{}c@{}}Stor- \\ age$\downarrow$\end{tabular}} & \multicolumn{5}{c|}{Depth}                  & \multicolumn{5}{c}{Intensity}               \\ \cline{2-3} \cline{6-15} 
		& CD$\downarrow$             & F-score$\uparrow$         &                           &                          & RMSE$\downarrow$   & MedAE$\downarrow$  & LPIPS$\downarrow$  & SSIM$\uparrow$   & PSNR$\uparrow$    & RMSE$\downarrow$   & MedAE$\downarrow$  & LPIPS$\downarrow$  & SSIM$\uparrow$   & PSNR$\uparrow$    \\ \hline
		DyNFL \cite{wu2024dynamic}                  & \underline{0.1357}         & \underline{0.9292}          & \underline{16.5\,s}                     & 19\,G                      & 7.4389 & 0.0336 & 0.3816 & 0.5391 & 20.7192 & 0.0663 & 0.0113 & 0.2410 & 0.6740 & 23.7657 \\
		LiDAR4D \cite{zheng2024lidar4d}                & 0.1505         & 0.9106          & 20.4\,s                     & \underline{9\,G}                       & \underline{4.9184} & \underline{0.0217} & \textbf{0.1670} & \textbf{0.7841} & \underline{24.3825} & \underline{0.0546} & \textbf{0.0077} & \underline{0.1456} & \textbf{0.7951} & \underline{25.4279} \\
		Ours                    & \textbf{0.0802}         & \textbf{0.9467}          & \textbf{0.9\,s}                      & \textbf{4.4\,G}                     & \textbf{4.7258} & \textbf{0.0180} & \underline{0.1696} & \underline{0.7744} & \textbf{24.7770} & \textbf{0.0531} & \underline{0.0079} & \textbf{0.1172} & \underline{0.7810} & \textbf{25.7017} \\ \hline
	\end{tabular}
	\caption{Quantitative comparison with SOTAs for rendering LiDAR data. We bold the best results and underline the second best.}
	\label{tab:1}
\end{table*}

\begin{figure*}[ht]
	\centering
	\includegraphics[width=0.9\linewidth]{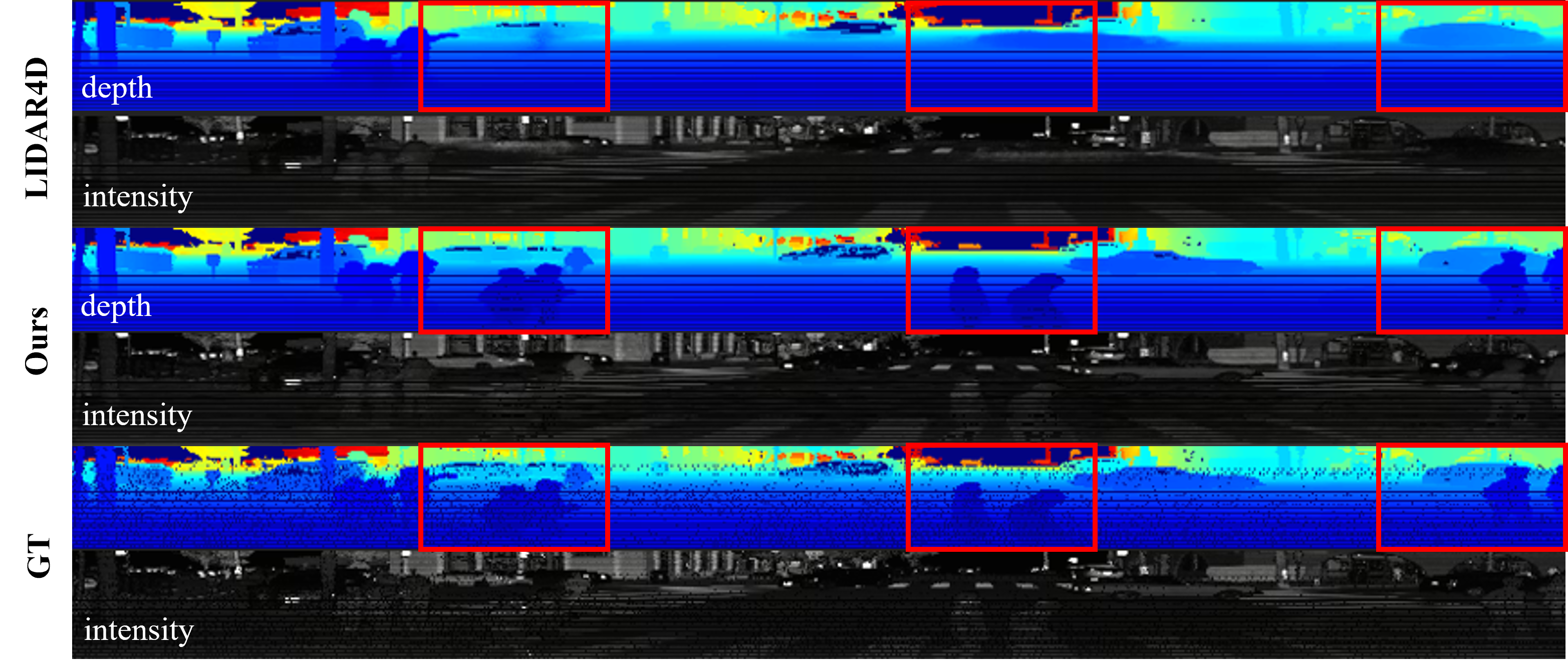}
	
	\caption{Qualitative comparison for LiDAR depth and intensity simulation.
	}
	\label{fig:range}
\end{figure*}

\begin{figure*}[ht]
	\centering
	\includegraphics[width=1.0\linewidth]{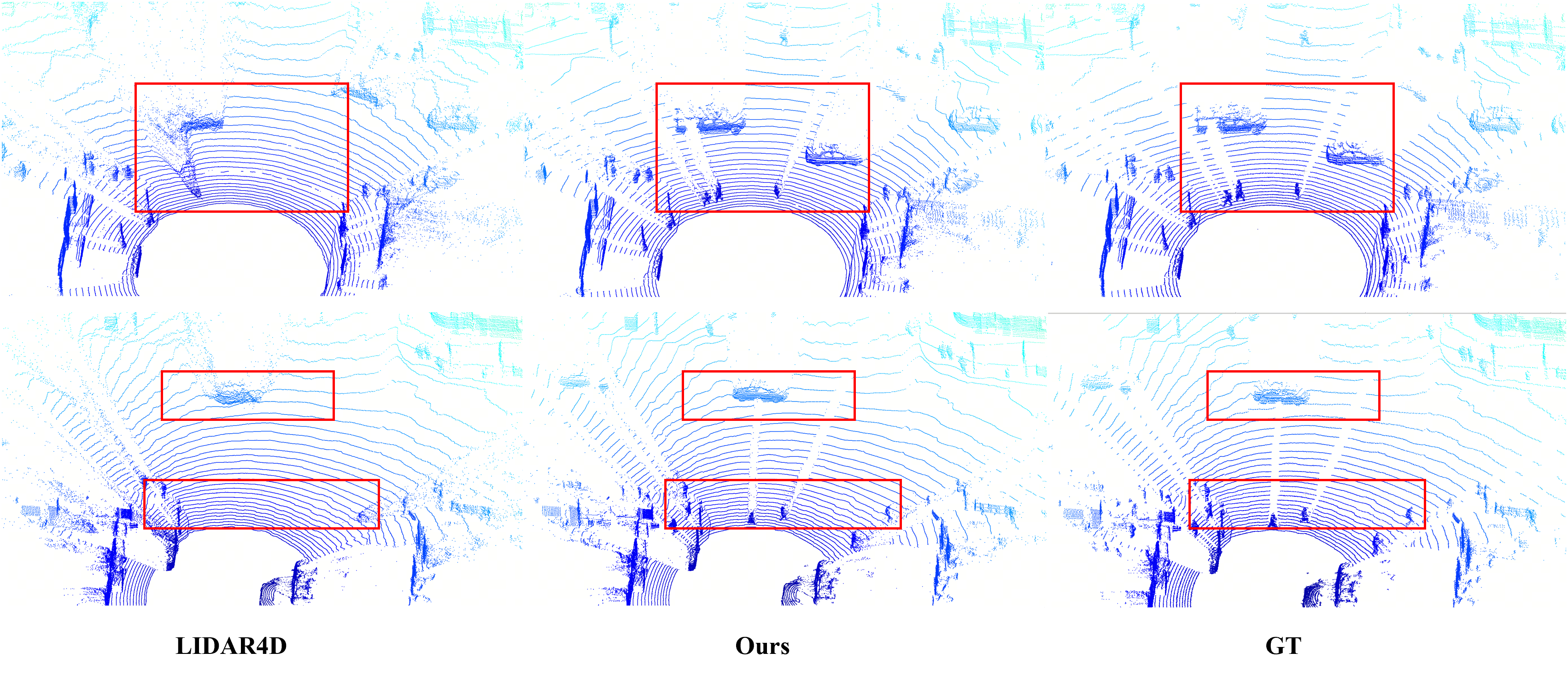}
	
	\caption{Qualitative comparison for LiDAR point simulation.
	}
	\label{fig:point}
\end{figure*}

We evaluated our Uni-Gaussians on the Waymo Open Dataset \cite{sun2020scalability}, which includes driving logs in real-world. We follow OmniRe \cite{chen2024omnire} to select 8 highly complex dynamic scenarios, which includes both vehicles, pedestrians and cyclists.

For the evaluation of LiDAR data rendering, we employ a comprehensive suite of metrics to assess the performance. Specifically, the Chamfer Distance \cite{fan2017point} quantifies the 3D geometric discrepancy between the generated and ground-truth point clouds via nearest-neighbor matching, and we also report the F-score with an error threshold set at 5 cm. Additionally, we introduce the Root Mean Square Error (RMSE) and Median Absolute Error (MedAE) to evaluate the pixel-wise error in the projected range images. To further assess the overall similarity, we utilize the Learned Perceptual Image Patch Similarity (LPIPS) \cite{zhang2018unreasonable}, Structural Similarity Index (SSIM) \cite{wang2004image}, and Peak Signal-to-Noise Ratio (PSNR). We evaluate both the depth and intensity reconstruction results, providing a holistic assessment of the rendering quality. For the evaluation of camera data rendering, we report PSNR and SSIM scores for full images.

For training, we utilize data from one LiDAR and five cameras with three forward-facing cameras., while the LiDAR range view is resized to a resolution of $64\times2650$ and the camera image is resized to a resolution of $640\times960$. During training, the weights for the losses in Sec. \ref{sec:opt} are as follows: $\lambda_r = 0.2$, $\lambda_{\text {depth}}=1$, $\lambda_{\text{intensity}} = 1$, $\lambda_{\text{raydrop}} = 0.5$, $\lambda_{\text{norm}} = 0.0001$. All experiments were conducted on a single NVIDIA-L4 GPU.

\subsection{Comparison of the SOTAs for Rendering Point Cloud}
\label{sec:exp_lidar}

We compare our Uni-Gaussians with two state-of-the-art methods for rendering LiDAR data, i.e., DyNFL \cite{wu2024dynamic} and LiDAR4D \cite{zheng2024lidar4d}. For all methods, we use their offfcial code.

The quantitative comparison is displayed in Table \ref{tab:1}. Our proposed Uni-Gaussians exhibits remarkable performance across all metrics in comparison to prior SOTA methods, demonstrating the superiority of camera and LiDAR simulation in dynamic reconstruction. In comparison to DyNFL and LiDAR4D, our approach has led to 40.9$\%$ and 46.7$\%$ reduction in the CD error of the novel-view point cloud synthesis respectively.  As shown in Fig. \ref{fig:range} and Fig. \ref{fig:point}, our method is capable of accurately and finely simulating various types of movable entities within a dynamic driving scenarios, whereas the latest SOTA LiDAR4D is unable to achieve this. This is attributed to the detailed modeling of scene graph. On the other hand, 2D Gaussian provides the compact scene representation which can utilize computational resources highly efficiently. In contrast, LiDAR4D is a NeRF-based method, where many of the rendered sampling points are located in the air. The computational overhead caused by rendering these air sampling points is essentially unnecessary. Consequently, when computational resources are limited, the rendering performance of LiDAR4D is suboptimal due to its low resource utilization. This further substantiates the superiority of our solution in the field of LiDAR simulation.

\subsection{Comparison of the SOTAs for Rendering Image}
\label{sec:exp_camera}

\begin{table}[]
	\centering
	\fontsize{7pt}{10pt}\selectfont
	\begin{tabular}{p{2.0cm}<{\centering}|p{1.5cm}<{\centering}p{1.5cm}<{\centering}}
		\hline
		Method   & SSIM$\uparrow$             & PSNR$\uparrow$            \\ \hline
		PVG \cite{chen2023periodic}      & 0.8780          & 28.6433          \\
		StreetGS \cite{yan2024street} & 0.8912          & 27.2023          \\
		OmniRe \cite{chen2024omnire}   & \textbf{0.9040} & \textbf{30.1287} \\
		Ours     & \underline{0.9013}    & \underline{29.6202}    \\ \hline
	\end{tabular}
	\caption{Quantitative comparison with SOTAs for rendering camera data on novel views. We bold the best results and underline the second best.}
	\label{tab:2}
\end{table}

\begin{figure*}[ht]
	\centering
	\includegraphics[width=0.9\linewidth]{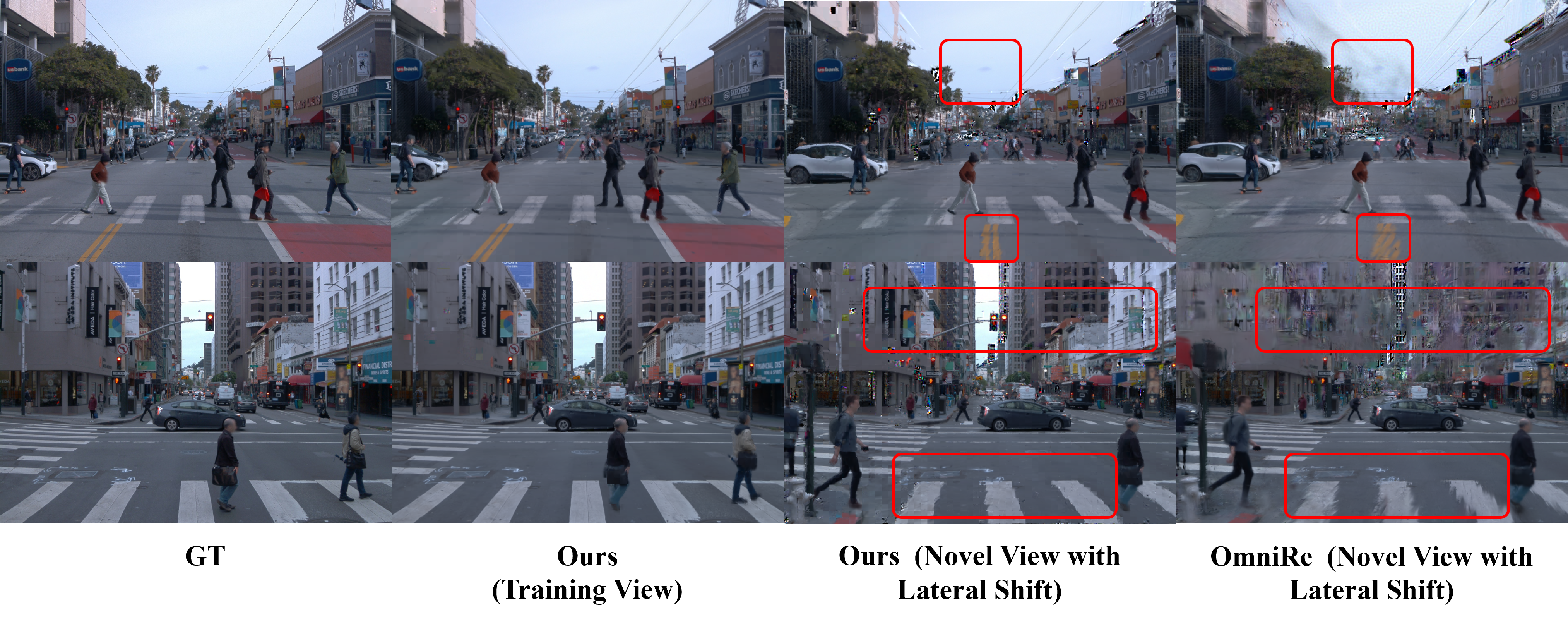}
	
	\caption{Qualitative comparison the synthesis of novel view with lateral shifts.
	}
	\label{fig:image}
\end{figure*}

\begin{table*}[]
	\centering
	\fontsize{7pt}{10pt}\selectfont
	\begin{tabular}{c|p{1.2cm}<{\centering}p{1.2cm}<{\centering}|p{1.2cm}<{\centering}p{1.2cm}<{\centering}p{1.2cm}<{\centering}p{1.2cm}<{\centering}p{1.2cm}<{\centering}}
		\hline
		\multirow{2}{*}{Method} & \multicolumn{2}{c|}{Point Cloud}  & \multicolumn{5}{c}{Depth}                                                                \\ \cline{2-8} 
		& CD$\downarrow$              & F-score$\uparrow$         & RMSE$\downarrow$            & MedAE$\downarrow$           & LPIPS$\downarrow$           & SSIM$\uparrow$            & PSNR$\uparrow$             \\ \hline
		Rasterization for Point Cloud           & 0.4154          & 0.8371          & 12.1743         & 0.4549          & 0.5993          & 0.1763          & 16.3759          \\
		Ray-Tracing for Point Cloud    & \textbf{0.0802} & \textbf{0.9467} & \textbf{4.7258} & \textbf{0.0180} & \textbf{0.1696} & \textbf{0.7744} & \textbf{24.7770} \\ \hline
	\end{tabular}
	\caption{Quantitative comparison with different approaches for point cloud rendering. We bold the best results.}
	\label{tab:4}
\end{table*}

\begin{figure}[ht]
	\centering
	\includegraphics[width=1.0\linewidth]{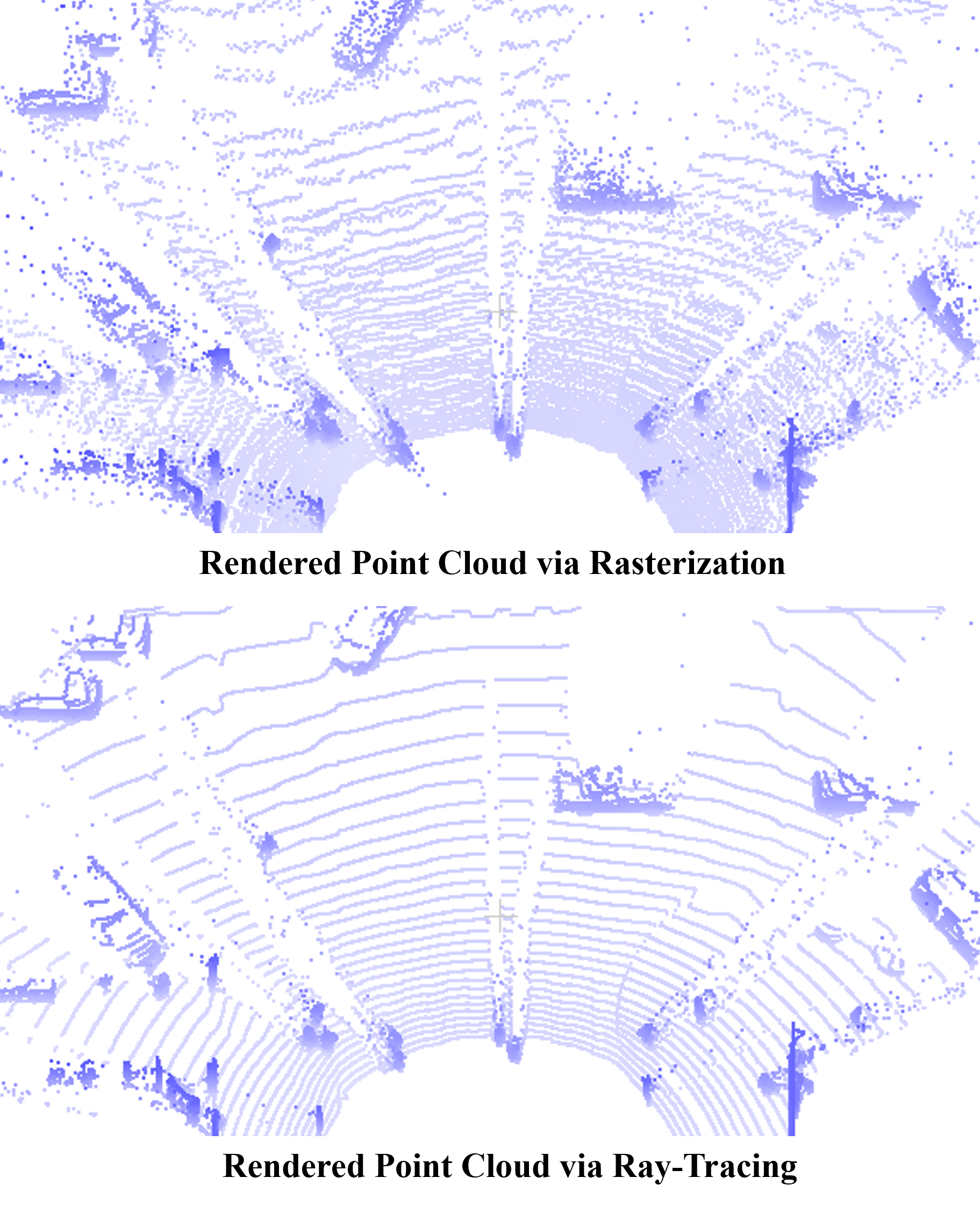}
	
	\caption{Visualized comparison between rasterization and ray-tracing for point cloud simulation.
	}
	\label{fig:ablation2}
\end{figure}

We compare our Uni-Gaussians with PVG \cite{chen2023periodic}, StreetGS \cite{yan2024street} and OmniRe \cite{chen2024omnire} to demonstrate that the unifying camera and LiDAR rendering with 2D Gaussians does not compromise the appearance quality compared to camera rendering with 3D Gaussians but also exhibits better generalization for novel view synthesis. For all methods, we use their offfcial code.

The quantitative comparison is displayed in Table \ref{tab:2}. Our Uni-Gaussians achieves comparable appearance performance with OmniRe and outperforms other methods on both PSNR and SSIM. Additionally, owing to the adoption of 2D Caussian which is a geometry-focused scene representation, our method exhibits superior generalization for novel view synthesis (as illustrated in Fig. \ref{fig:image}). In contrast, OmniRe fails to accurately simulate new views involving lateral shifts. This highlights the unique advantages of our approach.

\subsection{Ablation Study}
\label{sec:ablation}

This section aims to demonstrate that, the rendering approaches adopted for point cloud and image are indeed optimal for unifying Gaussian scenarios. To this end, we first validate that ray-tracing is the optimal rendering method for point cloud, followed by the confirmation that rasterization is the optimal rendering method for image data.

\subsection{Impact of Rendering Approaches for Point Cloud}
\label{sec:ablation2}

Although rasterization can generate depth information, experimental data indicate that the accuracy of such depth is not high. In contrast, ray-tracing is inherently more aligned with the active nature of LiDAR sensors which rely on the emission and reception of laser beams.

Results in Table \ref{tab:4} demonstrate that ray-tracing significantly outperforms rasterization in terms of rendering quality for point cloud data. Additionally, the visualization in Fig. \ref{fig:ablation2} indicates that rasterization tends to produce noticeable point positional disorder due to the inconsistency between the rasterization imaging process and the active nature of LiDAR sensors.

\subsection{Impact of Rendering Approaches for Image}
\label{sec:ablation1}

\begin{table}[]
	\centering
	\fontsize{7pt}{10pt}\selectfont
	\begin{tabular}{c|cc}
		\hline
		& Ray-Tracing for Image & Rasterization for Image \\ \hline
		Training & 20\,h                & 5\,h         \\
		Inference & 0.667\,s              & 0.008\,s    \\ \hline
	\end{tabular}
	\caption{Run time comparison with different approaches for image rendering.}
	\label{tab:3}
\end{table}

As mentioned in Sec. \ref{sec:method}, both rasterization and ray-tracing can ensure satisfactory simulation outcomes for image data. Given that rasterization offers faster rendering speeds, we utilize rasterization to render image data.

Results in Table \ref{tab:3} indicate that the inference speed of rasterization for single image data is 83X faster than that of Gaussian ray tracing, which significantly affects the efficiency of both training and inference processes. This further substantiates the rationality of our decision to employ rasterization for rendering images.

\section{Conclusion}
\label{sec:conclusion}

This paper introduces a novel and unifying camera and LiDAR simulation framework leveraging 2D Gaussians, specifically designed to deliver efficient and accurate simulation for dynamic driving scenarios. We advocate the use of a unified Gaussian representation to harmonize the simulation process, while employing distinct rendering approaches tailored to the unique characteristics of different data types. For image data, rasterization is chosen as the rendering method, capitalizing on its significantly faster rendering speed compared to ray-tracing, thereby ensuring real-time performance and computational efficiency. Conversely, for point cloud data, ray-tracing is utilized to address the challenges posed by the inconsistent imaging process of rasterization and LiDAR sensors. Extensive experimental results demonstrate that our proposed method outperforms existing solutions, establishing itself as the optimal approach for achieving high-fidelity camera and LiDAR simulation in dynamic driving scenarios.

\bibliographystyle{IEEEtrans}
\bibliography{IEEEabrv,IEEEExample}

\end{document}